\documentclass[10pt,twocolumn,letterpaper]{article}

\usepackage{cvpr}              

\usepackage{graphicx}
\usepackage{amsmath}
\usepackage{amssymb}
\usepackage{booktabs}
\usepackage[accsupp]{axessibility}

\usepackage[pagebackref=false, breaklinks=true, letterpaper=true, colorlinks, urlcolor=gray,
            citecolor=citecolor, linkcolor=linkcolor, bookmarks=false]{hyperref}

\usepackage[capitalize]{cleveref}
\crefname{section}{Sec.}{Secs.}
\Crefname{section}{Section}{Sections}
\Crefname{table}{Table}{Tables}
\crefname{table}{Tab.}{Tabs.}

\usepackage{bm}
\usepackage[british,english,american]{babel}
\usepackage{microtype}      
\usepackage{enumitem}
\usepackage{lipsum}
\usepackage{multirow}
\usepackage{xcolor}
\usepackage{colortbl}

\usepackage{algorithm}
\usepackage[noend]{algpseudocode}
\usepackage{algorithmicx}

\usepackage{etoolbox}
\makeatletter
\AfterEndEnvironment{algorithm}{\let\@algcomment\relax}
\AtEndEnvironment{algorithm}{\kern2pt\hrule\relax\vskip3pt\@algcomment}
\let\@algcomment\relax
\newcommand\algcomment[1]{\def\@algcomment{\footnotesize#1}}
\renewcommand\fs@ruled{\def\@fs@cfont{\bfseries}\let\@fs@capt\floatc@ruled
  \def\@fs@pre{\hrule height.8pt depth0pt \kern2pt}%
  \def\@fs@post{}%
  \def\@fs@mid{\kern2pt\hrule\kern2pt}%
  \let\@fs@iftopcapt\iftrue}
\makeatother

\newlength\savewidth\newcommand\shline{\noalign{\global\savewidth\arrayrulewidth
  \global\arrayrulewidth 1pt}\hline\noalign{\global\arrayrulewidth\savewidth}}
\newcommand{\tablestyle}[2]{\setlength{\tabcolsep}{#1}\renewcommand{\arraystretch}{#2}\centering\footnotesize}

\newcommand{\I}[1]{\mathbf{I}_{#1}}
\newcommand{\St}[1]{\mathbf{S}_{#1}}
\newcommand{\mpara}[2]{\vspace{#1}\noindent\textbf{#2}}
\newcommand{\best}[1]{\textbf{#1}}

\def\up#1{(\textcolor[rgb]{0,0.75,0.25}{$\uparrow${#1}})}

\newcommand{\inc}[1]{\textcolor[rgb]{0,0.75,0.25}{$\uparrow${#1}}}
\newcommand{\dec}[1]{\textcolor[rgb]{1,0,0}{$\downarrow${#1}}}

\definecolor{create}{RGB}{46, 117, 182}
\definecolor{remove}{RGB}{197, 90, 17}

\newcommand{\Create}{\textcolor{create}{\bf C}}
\newcommand{\Remove}{\textcolor{remove}{\bf R}}

\definecolor{citecolor}{HTML}{0071BC}
\definecolor{linkcolor}{HTML}{ED1C24}

\definecolor{convcolor}{HTML}{412F8A}
\definecolor{vitcolor}{HTML}{fc8e62}

\newcommand{\convcolor}[1]{\textcolor{convcolor}{#1}}
\newcommand{\vitcolor}[1]{\textcolor{vitcolor}{#1}}
\newcommand{\vb}{\vitcolor{$\mathbf{\circ}$\,}}
\newcommand{\cb}{\convcolor{$\bullet$\,}}
\newcommand{\gr}{\rowcolor[gray]{.95}}

\newcommand\blfootnote[1]{%
\begingroup
\renewcommand\thefootnote{}\footnote{#1}%
\addtocounter{footnote}{-1}%
\endgroup
}

\def\cvprPaperID{5577} 
\def\confName{ICCV}
\def\confYear{2023}

\begin{document}

\title{
\vspace{-1mm}
Scalable Multi-Temporal Remote Sensing Change Data Generation\\ via Simulating Stochastic Change Process
\vspace{-5mm}
}

\author{
Zhuo Zheng$^{1,2}$,
Shiqi Tian$^{1}$,
Ailong Ma$^{1}$,
Liangpei Zhang$^{1}$,
Yanfei Zhong$^{1, *}$ \\
$^1$Wuhan University~~$^2$Stanford University\\
{\small \hypersetup{urlcolor=magenta}\url{https://github.com/Z-Zheng/Changen}}
\vspace{-2mm}
}

\twocolumn[{%
      \renewcommand\twocolumn[1][]{#1}%
      \maketitle
      \begin{center}
        \noindent
        \vspace*{-2em}
        \includegraphics[width=\textwidth]{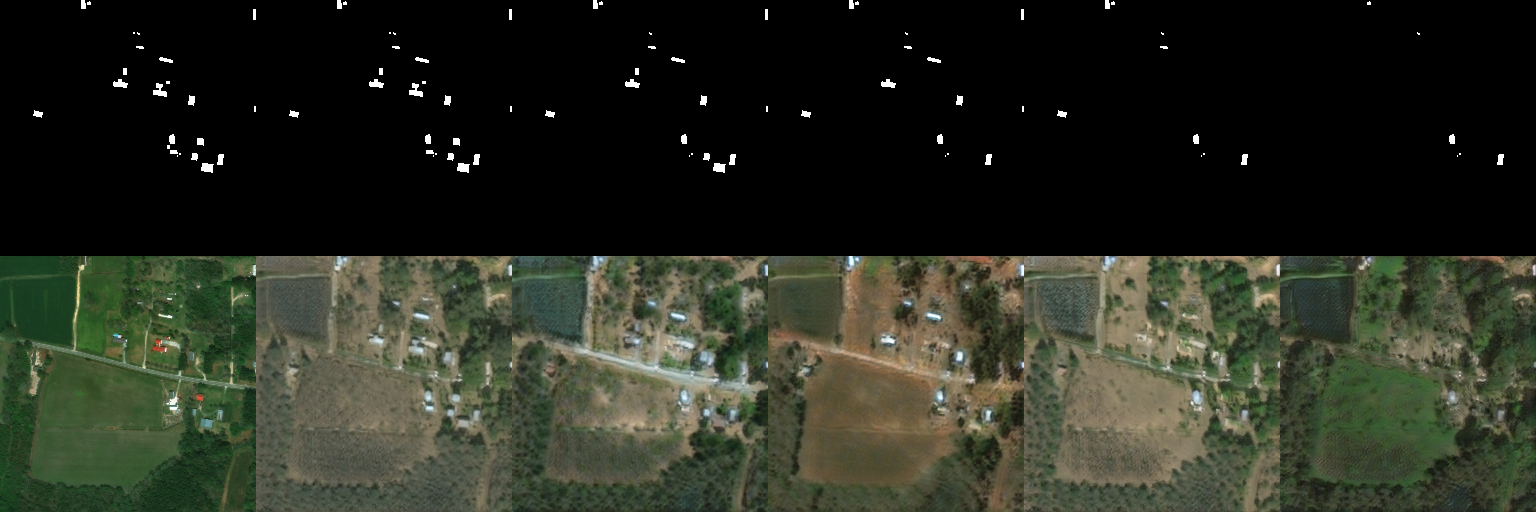}
        \vspace*{-6mm}
        \captionsetup{type=figure}
        \caption{This village is disappearing, which is a stochastic change process simulated by our generative probabilistic change model (GPCM).
        This is also synthetic multi-temporal change data produced by our change generator (Changen), a GAN-based GPCM, enabling object change data generation with controllable object property (\eg, scale, position, orientation), and change event.
        }
        \vspace*{-1mm}
        \label{figs:teaser}
      \end{center}
    }]

\begin{abstract}
  \vspace{-1.8mm}
  Understanding the temporal dynamics of Earth's surface is a mission of multi-temporal remote sensing image analysis, significantly promoted by deep vision models with its fuel---labeled multi-temporal images.
  However, collecting, preprocessing, and annotating multi-temporal remote sensing images at scale is non-trivial since it is expensive and knowledge-intensive.
  In this paper, we present a scalable multi-temporal remote sensing change data generator via generative modeling, which is cheap and automatic, alleviating these problems.
  Our main idea is to simulate a stochastic change process over time.
  We consider the stochastic change process as a probabilistic semantic state transition, namely generative probabilistic change model (GPCM), which decouples the complex simulation problem into two more trackable sub-problems, \ie, change event simulation and semantic change synthesis.
  To solve these two problems, we present the change generator (Changen), a GAN-based GPCM, enabling controllable object change data generation, including customizable object property, and change event.
  The extensive experiments suggest that our Changen has superior generation capability, and the change detectors with Changen pre-training exhibit excellent transferability to real-world change datasets.
\end{abstract}

\vspace{-8mm}
\blfootnote{* corresponding author (zhongyanfei@whu.edu.cn)}
\section{Introduction}
\label{sec:intro}
Change detection is one of the most fundamental Earth vision tasks to understand the temporal dynamics of Earth's surface.
Tremendous progress in change detection has been achieved by joint efforts of remote sensing and computer vision communities.
Deep change detection models \cite{daudt2018fully, changestar, changeos, zheng2022changemask} represented by Siamese networks \cite{siamese} have dominated in recent years.
The key behind their success lies in large-scale training datasets \cite{levircd, shen2021s2looking, DEN, hiucd}.
However, building a large-scale remote sensing change detection dataset is difficult and expensive,
because collecting, preprocessing, annotating remote sensing images needs more expertise and efforts.

Synthetic data, as an alternative, is a promising direction to alleviate the data hungry.
There are currently two main paradigms for change detection data synthesis in the remote sensing domain, \ie, graphics-based \cite{bourdis2011constrained, kolos2019procedural} and data augmentation-based \cite{chen2021adversarial} approaches.
The graphics-based methods synthesize images by rendering manual constructed 3D models.
IAug \cite{chen2021adversarial}, as a data augmentation-based approach, synthesizes new image pairs by pasting object instances into existing bitemporal image pairs.

However, the scalability and diversity of these conventional synthetic change datasets remain limited.
This is because 3D modeling and rendering need expertise and tremendous efforts for graphics-based approaches, and an existing change dataset is needed for data augmentation-based approaches.
Besides, the synthetic data is aimed to assist the change detection model in improving the performance on real-world data. 
However, the relationship between the quality of synthetic data and the transferability of features learned from synthetic data is still unclear due to scale-limited data of graphics-based approaches and the coupling of synthetic and real-world data of data-augmentation based approaches.

In this paper, we present a scalable multi-temporal change data generator via generative modeling.
Our data generator is aimed to generate realistic and diverse multi-temporal labeled images from a single-temporal image and its semantic segmentation mask, by \textit{simulating the change process}.

\begin{figure}[ht]
  \centering
   \includegraphics[width=0.9\linewidth]{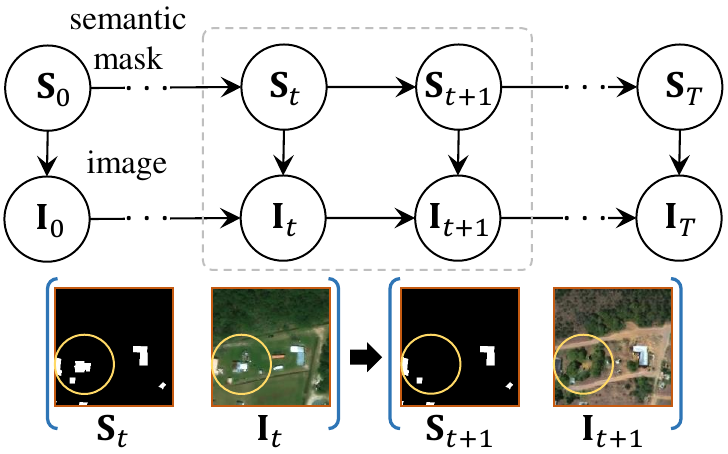}
   \caption{Generative Probabilistic Change Model.
   }
   \label{fig:gpcm}
\end{figure}

To this end, we first describe the stochastic change process as a probabilistic graphical model, namely \textit{generative probabilistic change model} (GPCM), considering each image and semantic mask as random variables, as shown in Fig.~\ref{fig:gpcm}.
The change is always driven by the event, therefore, we make a Markov assumption that the image and its semantics at time $t+1$ only depends on the image and its semantics at time $t$, to simplify this modeling problem.
Based on this condition, the whole problem can be decoupled to two sub-problems, \ie, the change event simulation at semantic-level and the semantic change synthesis at image-level.

To solve above two sub-problems, we propose a generative model called \textit{Changen}, which is a GPCM parameterized with generative adversarial networks (GANs).
Our Changen creates or removes objects in the semantic mask at time $t$, as a stochastic change event, to generate a new semantic mask at time $t+1$, and synthesizes a post-event image at time $t+1$, by progressively applying simulated semantic changes to 
the pre-event image at time $t$.

As demonstration, our Changen generates a large-scale building change detection dataset with diverse object properties (e.g., scale, position, orientation) and two change events.
The change detector pre-trained on this dataset has the superior transferability on the real-world building change detection datasets, significantly outperforming commonly used ImageNet \cite{imagenet} pre-training, and additionally possessing zero-shot prediction capability.
More importantly, based on our Changen, we find that the temporal diversity of synthetic change data is a key factor in ensuring transferability after model pre-training.
The main contributions of this paper are summarized as follows:
\vspace{-3mm}
\begin{itemize}[leftmargin=*]
\item \textbf{Generative change modeling} decouples the complex stochastic change process simulation to more tractable change event simulation and semantic change synthesis. 
\vspace{-3mm}
\item \textbf{Change generator}, \ie, Changen, enables object change generation with controllable object property (\eg, scale, position, orientation), and change event.
\vspace{-3mm}
\item \textbf{Our synthetic change data pre-training} empowers the change detectors with better transferability and zero-shot prediction capability.
 
\end{itemize}

\section{Related Work}
\label{sec:relwork}

\mpara{0em}{Change Data Synthesis.}
Based on the computer graphics, the early studies mainly use game engines to generate realistic remote sensing images from existing assets.
For example, AICD dataset~\cite{bourdis2011constrained}, consists of 1,000 pairs of 800$\times$600 images that are automatically generated by the built-in rendering engine of a computer game.
However, due to limited assets and an underdeveloped rendering engine, this dataset is of low diversity and graphics quality.
To improve the diversity and realism of synthesized images, a semi-automatic data generation pipeline \cite{kolos2019procedural} is proposed to synthesize a change detection dataset, which adopts cartographic data for manual 3D scene modeling and the professional game engine for rendering.
Another way is based on data augmentation, especially Copy-Paste \cite{copypaste}, \eg, IAug \cite{chen2021adversarial} adopts SPADE \cite{spade} to generate object instances and randomly pastes them into an existing change detection dataset to increase training samples.
Our work provides a new perspective, \ie, generative modeling for change detection data synthesis.
Benefiting from a deep generative model, our dataset generator is automatic and relies only on mono-temporal data, ensuring scalability and economy of synthetic data.

\begin{figure*}[ht]
  \centering
   \includegraphics[width=1.\linewidth]{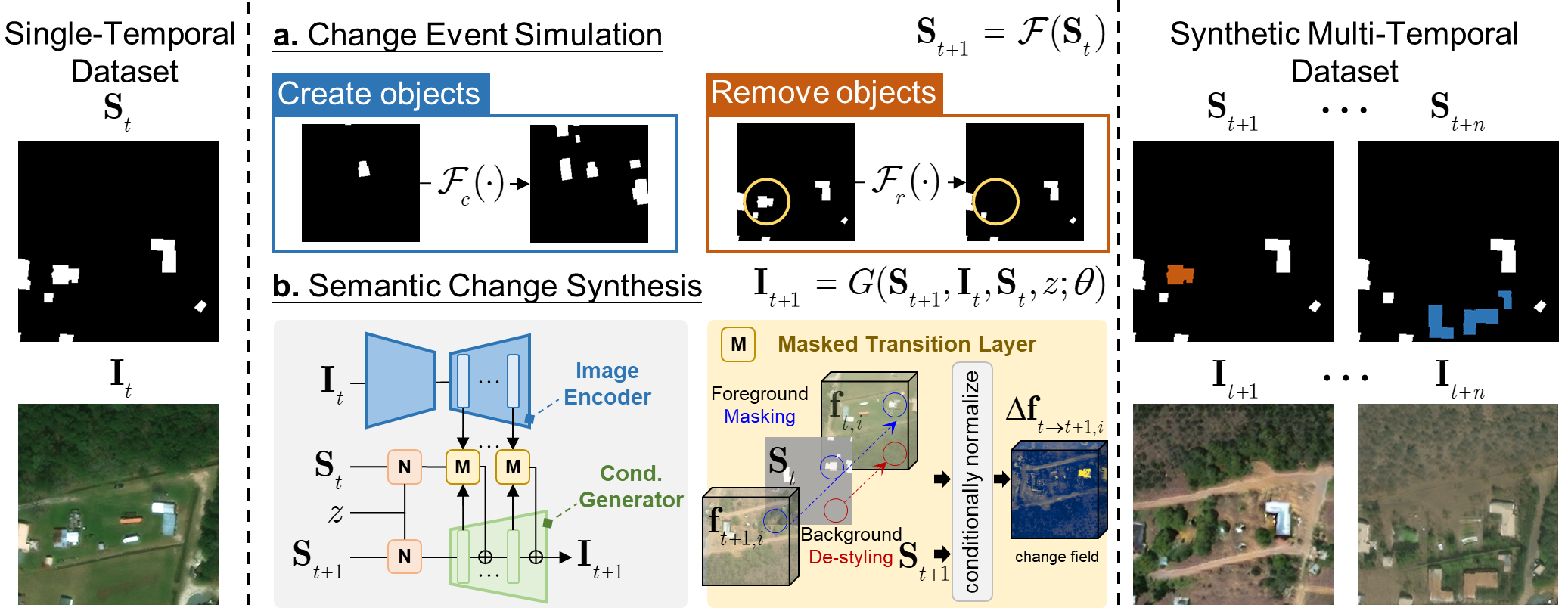}
   \caption{\textbf{Our Changen architecture}.
   The change event simulation enables adding or removing objects in the semantic mask at time $t$ to customize new semantic masks at times $t+1~{\rm to}~n$.
   For the semantic change synthesis, the new images at times $t+1~{\rm to}~n$ will be synthesized by progressively accumulating estimated change fields on the image at time $t$.
   Our Changen can generate the multi-temporal dataset with controllable scene layout, object property (\eg, scale, position, orientation, see $\I{t+n}$), and change event.
   Legend: \textcolor{create}{\textbf{Create}}; \textcolor{remove}{\textbf{Remove}}.
   }
   \label{fig:changen}
\end{figure*}

\mpara{0.em}{Semantic Image Synthesis.}
Recent studies \cite{pix2pixHD, spade, zhu2020sean, tan2021diverse, shi2022retrieval, oasis} mainly focus on translating the semantic segmentation label to the image based on different conditional GANs \cite{cGAN}.
pix2pix \cite{pix2pix} and pix2pixHD \cite{pix2pixHD}, as early representative methods, directly use semantic mask to generate the image.
SPADE \cite{spade} reveals that normalization layers tend to ``wash away'' semantic information, and propose a spatially adaptive conditional normalization layer to incorporate semantic information and spatial layout.
The following studies further improve SPADE on region-level control ability \cite{zhu2020sean}, diversity \cite{tan2021diverse}, better trade-off between fidelity and diversity \cite{shi2022retrieval}.
These methods strongly depend on VGG-based perceptual loss \cite{dosovitskiy2016generating, johnson2016perceptual} to guarantee image quality, however, this increases the complexity of the whole pipeline.
To simplify GAN models for semantic image synthesis, OASIS \cite{oasis} reveals that the segmentation-based discriminator is the key to enable GAN model to synthesize high-quality images with only adversarial supervision.
Our work extends conventional semantic image synthesis in temporal dimensionality, enabling semantic time-series synthesis from single-temporal data and with better consistency on the spatial layout.

\section{Generative Probabilistic Change Model}
\label{sec:method}
The main idea of our dataset generator is to simulate stochastic change process started from each single-temporal image and its semantic mask.
To this end, we frame the stochastic change process as a probabilistic graphical model shown in Fig.~\ref{fig:gpcm}, to describe the relationship between the distributions of variables (\ie, images, semantic masks) over time.
In this way, we can parameterize several smaller factors instead of directly parameterizing high-dimensionality joint distribution over all variables, simplifying this simulation problem.
Based on this modeling framework, we provide a parameterization with a deep generative model, to synthesize the multi-temporal change detection dataset from single-temporal data.

\subsection{Generative Change Modeling Framework}
The image $\I{t}$ has a pre-defined pixel-wise semantics $\St{t}$ at time $t$.
Given a stochastic change event occurred in the time period of $t$ to $t+1$, the image $\I{t}$ gradually evolves into the image $\I{t+1}$ with a semantic transition from $\St{t}$ to $\St{t+1}$.
The joint distribution of this stochastic change process is denoted as $P_{\rm scp}:=P(\St{t+1}, \St{t}, \I{t+1}, \I{t})$.
Here, we make a Markov assumption that the post-event image-semantics pair ($\I{t+1}, \St{t+1}$) only depends on the pre-event image-semantics pair ($\I{t}, \St{t}$), because the change is always event-driven.
Based on this assumption, $P_{\rm scp}$ can be factorized as $P(\St{t+1}, \I{t+1}| \St{t}, \I{t})P(\St{t}, \I{t})$.
Since the simulation is a generative task, \ie, semantics-to-images, the graph structure of the set of these four random variables can be described as shown in Fig.~\ref{fig:gpcm}.
The following factorization can be obtained:
\begin{equation}
   P_{\rm scp} = P(\I{t+1}|\St{t+1}, \I{t})P(\St{t+1}|\St{t})P(\I{t}| \St{t})P(\St{t})
\end{equation}
where the semantics distribution $P(\St{t})$ and corresponding conditional image distribution $P(\I{t}| \St{t})$ can be approximately seen as two known distributions because the single-temporal data is given for this simulation problem.
This means that the semantic transition distribution $P(\St{t+1}|\St{t})$ and the conditional image distribution $P(\I{t+1}|\St{t+1}, \I{t})$ need to be further estimated, so as to sample from $P_{\rm scp}$.

Through above modeling, the whole simulation problem can be decoupled to two sub-problems, \ie, the \textit{change event simulation} to approximate $P(\St{t+1}|\St{t})$ and the \textit{semantic change synthesis} to approximate $P(\I{t+1}|\St{t+1}, \I{t})$.

\subsection{Changen: GPCM parameterized with GANs}
To solve above two sub-problems, we present a parameterization for GPCM with the learning-free function $\mathcal{F}(\cdot)$ for change event simulation and the well-designed deep generative model $G$ for semantic change synthesis, as follows:
\vspace{-1.5em}
\begin{align}
  & \St{t+1} = \mathcal{F}(\St{t})   \\
  & \I{t+1}  = G(\St{t+1}, \I{t}, \St{t}, z; \theta), z\sim \mathcal{N}(\mathbf{0}, \mathbf{1})
\end{align}
Fig.~\ref{fig:changen} illustrates the whole architecture, introduced next.

\subsubsection{Change Event Simulation}
Sampling from $P(\St{t+1}|\St{t})$ is to obtain $\St{t+1}$ as a semantic guidance for the subsequent semantic change synthesis.
We consider two common change events, \ie, constructing new objects and destroying existing objects in real-world.
From the perspective of image editing, we refer to these two events as creating objects and removing objects.
To simulate these two events in semantic level, we design two learning-free functions, \ie, $\mathcal{F}_c(\cdot): R^{h\times w}\rightarrow R^{h\times w}$ for object creation and $\mathcal{F}_r(\cdot): R^{h\times w}\rightarrow R^{h\times w}$ for object removal.
In detail, we first randomly select some instances from semantic segmentation labels.
The selected instances are either pasted on the rest area of the semantic segmentation label to simulate object creation or directly assigned the background label to simulate object removal, as shown in Fig.~\ref{fig:changen}a.

\vspace{-1em}

\subsubsection{Semantic Change Synthesis}
Sampling from $P(\I{t+1}|\St{t+1}, \I{t})$ is to obtain the post-event image $\I{t+1}$.
To this ends, we design a deep generative model as shown in Fig.~\ref{fig:changen}b, which has an image encoder that maps the pre-event image to multi-scale latent representations and a conditional decoder that progressively generates the post-event image from the latent representations, the post-event semantic representation, and Gaussian noise. 
Besides, the masked transition layers inserted between the image encoder and the conditional decoder, is to reassemble their feature maps for avoiding mode collapse caused by a trivial solution.

\mpara{0em}{Image Encoder.}
We adopt the U-Net \cite{unet} with ResNet-18 \cite{resnet} as the image encoder to obtain multi-scale latent representations $\{\mathbf{f}_{t, i}\}_{i=0}^5$ of the pre-event image, where $\mathbf{f}_{t, i}$ has a size of $\frac{512}{2^i}\times \frac{h}{2^{(5-i)}} \times \frac{w}{2^{(5-i)}}$, and $h, w$ are the image size.
Spectral normalization \cite{sn} is applied to all conv layers.

\mpara{0.5em}{Conditional Decoder.}
Our decoder is stacked by six residual blocks with spatially-adaptive group normalization \cite{spade, gn} for consistent training and inference.
The decoder takes one-hot encoded post-event semantic mask concatenated with a 3D Gaussian noise map \cite{oasis} in channel axis as input and yields multi-scale post-event feature maps $\{\mathbf{f}_{t+1, i}\}_{i=0}^5$ at time $t+1$.
The first feature map $\mathbf{f}_{t+1, 0}$ is obtained by computing a 3$\times$3 conv layer over 32$\times$ downsampled input.
The remaining post-event feature map $\mathbf{f}_{t+1, i+1}$ is computed as follows:
\begin{align}
  &\Delta \mathbf{f}_{t\rightarrow t+1, i} = \mathbf{M}_i(\mathbf{f}_{t, i}, \mathbf{f}_{t+1, i}, \St{t}, \St{t+1}) \\
  &\mathbf{f}_{t+1, i+1} = G_i(\mathbf{f}_{t+1, i} + \Delta \mathbf{f}_{t\rightarrow t+1, i}, \St{t+1}, z)
\end{align}
where the change field $\Delta \mathbf{f}_{t\rightarrow t+1}$ is first computed by the masked transition layer $\mathbf{M}$ (described next in detail).
The each spatial position of the change field is a change vector encoding change information from the pre-event image to the post-event image.
These change fields are progressively applied to intermediate feature maps of our decoder, and $i$-th result is fed to the residual block $G_i$ for further conditional synthesis.
The tanh activation is applied to the final feature map to obtain the synthesized post-event image.
To stable training, the spectral normalization is also applied in our decoder.
The layer $\mathbf{N}$ is the noise injection via \texttt{concat}.

\begin{algorithm}[t]
  \caption{Bitemporal Adversarial Learning\label{alg:bal}}
  \algcomment{\fontsize{7.2pt}{0em}\selectfont
  $\textbf{v}_g: = (z, \St{t+1},\I{t})$.
  For simplicity, we ignore the notation of the classification score converter for the discriminator $D$, which is identical to \cite{oasis}.
  }
  \algrenewcommand\algorithmicindent{.5em}
  \newcommand{\HS}{\hspace{-3em}}
  
  \begin{algorithmic}[1]
  \Require    
  \Statex $\mathcal{D}_{t}\sim P(\I{t}, \St{t})$: Single-temporal dataset
  \Statex $N$: Number of iterations
  \For{$n=1$ {\bfseries to} $N$}
  
    \HS$\triangleright$ random mini-batch sampling
    \State $\I{t}, \St{t}\sim \mathcal{D}_{t}$
  
    \HS$\triangleright$ stochastic change event simulation
    \State $\St{t+1} \gets \mathcal{F}(\St{t})$
  
    \HS$\triangleright$ semantic change synthesis
  
    \State $\widetilde{\mathbf{I}}_{t+1} \gets G(\St{t+1}, \I{t}, \St{t}, z; \theta), z\sim \mathcal{N}(\mathbf{0}, \mathbf{1})$
    
    \HS$\triangleright$ update the generator once
    \State $\min\limits_{\theta}\mathbb{E}_{\mathbf{v}_g}[D(\widetilde{\mathbf{I}}_{t+1}, \St{t+1};\phi)]$ 
    
    \HS$\triangleright$ update the discriminator once (\textcolor{cyan}{ours})
    \State $\max\limits_{\phi}\mathbb{E}_{\mathbf{v}_g}[D(\widetilde{\mathbf{I}}_{t+1}, \St{t+1};\phi)]$ + \textcolor{cyan}{$\mathbb{E}_{\I{t}}[D(\mathbf{I}_{t}, \St{t};\phi)]$}

    \HS$\triangleright$ update the discriminator once (\textcolor{pink}{classical}, unavailable)
  
    \hspace{-7.mm}$\max\limits_{\phi}\mathbb{E}_{\mathbf{v}_g}[D(\widetilde{\mathbf{I}}_{t+1}, \St{t+1};\phi)]$ + \textcolor{pink}{$\mathbb{E}_{\I{t+1}}[D(\mathbf{I}_{t+1}, \St{t+1};\phi)]$}

  \EndFor
  \State {\bfseries return} $\theta$
  \end{algorithmic}
  \end{algorithm}

\mpara{0.5em}{Masked Transition Layer.}
With the guidance of the pre-event image, the synthesis of the post-event image can be easier.
However, this also causes a trivial solution, \ie, the generative model directly copies the content from the pre-event image rather than synthesizing visual content, yielding completely incorrect change information.
To avoid this ``feature leakage'', we propose the masked transition layer, as illustrated in Fig.~\ref{fig:changen}b, to yield a change field to progressively update the post-event feature map rather than to produce it directly.

Our masked transition layer first masks\footnote{Pseudo code of \textit{Masking} in a PyTorch-like style:\\ ${\rm mask=}~(\St{t}~{\rm ==~fg})$; ${\rm return~torch.where}({\rm mask}, \mathbf{f}_{t+1, i}, \mathbf{f}_{t, i}$)} pre-event foreground region features with the post-event feature maps and de-styles\footnote{For simplicity, we adopt a 1$\times$1 conv followed by instance normalization \cite{in} and a Leaky ReLU for De-styling.} the pre-event background region features for explicitly weakening over-emphasized pre-event features.
We further inject the post-event semantic information into the weakened pre-event features via a spatially-adaptive normalization conditioned by the post-event semantic mask, yielding a change field.

\mpara{0.5em}{Bitemporal Adversarial Learning.}
Since there is no ground-truth for the image to be generated, we propose the bitemporal adversarial learning to train our Changen, as illustrated in Algorithm~\ref{alg:bal}.
Like all adversarial learning for semantic image synthesis, we have a conditional discriminator to determine whether the image-mask pair is real or fake.
Following OASIS \cite{oasis}, we adopt a segmentation-based discriminator $D$ with spectral normalization and train our Changen with only conditional adversarial losses.
Unlike classical conditional adversarial losses to optimize the discriminator, we use the pair ($\I{t}, \St{t}$)  as the real target instead of the pair ($\I{t+1}, \St{t+1}$) because there is no ground truth for the image $\I{t+1}$, as presented in the line 6 of  Algorithm~\ref{alg:bal}.

\begin{figure*}[ht]
  \centering
   \includegraphics[width=1.\linewidth]{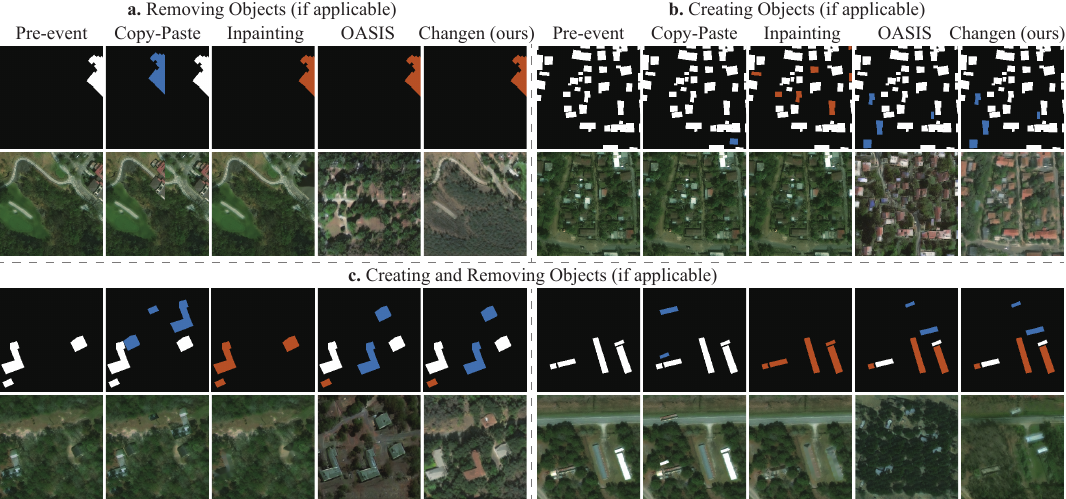}
   \caption{\textbf{Visual comparison of change data synthesis results}.
   Each pre-event column presents a real pre-event image-mask pair.
   Other columns present a synthesized post-event image-mask pair for each method.
   \textbf{a.} and \textbf{b.} simulate object removal and object creation, respectively.
   \textbf{c.} simulates more complex scenarios that include object removal and object creation simultaneously.
   Legend: \textcolor{create}{\textbf{Create}}; \textcolor{remove}{\textbf{Remove}}.
   }
   \label{fig:syn_all_m}
   \vspace{-1em}
\end{figure*}

\section{Experiments}
\label{sec:exp}

\subsection{Experimental Setting}

\mpara{-0.5em}{Single-temporal dataset}.
We use the pre-disaster part of xView2 dataset \cite{gupta2019xbd} as our single-temporal dataset, called \textit{xView2 pre-disaster}.
The \texttt{train} and \texttt{tier3} splits are used to train generative models.
The \texttt{hold} split, which is cropped into 256$\times$256 non-overlapped patches, is used to evaluate the quality of synthesized images.

\mpara{0.em}{Training generative models}.
We use Adam \cite{adam} with $\beta_1 = 0$ and $\beta_2=0.999$ as our optimizer for both the generator and the discriminator.
The learning rate is 0.0001 for the generator and 0.0004 for the discriminator.
We use a mini-batch size of 32 in 8 GPUs.
We train all generative models for 100k iterations with the same data and settings.
Random flip, rotation, transpose, scale jitter, cropping into 256$\times$256 are adopted for training data augmentation.

\mpara{0.em}{Change dataset generation}.
We sample pre-event image-mask pairs from the xView2 pre-disaster, resulting $\sim$90k pairs.
Changen is applied to these pre-event pairs for generating post-event pairs, resulting a change dataset.
Each pre-event mask can be mapped to $n\in \{1,2,...\}$ post-event masks, therefore, we refer to the change dataset with $n$ as \texttt{Changen-\{n*90\}k}, which means that this dataset consists of \texttt{n*90}k synthetic bitemporal training sample pairs.
We use \texttt{Changen-90k} as default for fast experiments.

\mpara{0.em}{Pre-training on synthetic dataset}.
We use ChangeStar \cite{changestar}, a simple multi-task change detection architecture, with an optimized configuration of FarSeg \cite{farseg,zheng2023farseg++} with 96 channels, ChangeMixin with $N=1, d_c=96$, and ResNet-18 backbone, which denotes ChangeStar (1$\times$96), as default for simplicity.
We pre-train the change detector for 20 epochs on each synthetic dataset if not specified.
SGD with a momentum of 0.9 and a weight decay of 0.0001 is used as our optimizer.
We use a mini-batch size of 64 with a initial learning rate of 0.03 and multiplied by $(1 - \frac{n}{N})^\gamma$ with $\gamma=0.9$.
Random flip, rotation, transpose, color jitter are adopted for training data augmentation.

\begin{table}[htb]
  \caption{Comparison with other change data generation approaches on the image quality.
    \label{tab:gen_model}}
  \centering
  \small
  \tablestyle{7pt}{1.2}
  \begin{tabular}{l|l|cc}
    Method              &  Modeling   & FID$\downarrow$ & IS$\uparrow$  \\
    \shline
    SPADE \cite{spade} + GPCM  & \multirow{2}{*}{$P(\I{t}|\St{t})$}        & 204.01    &  3.41     \\
    OASIS \cite{oasis} + GPCM  &                                               &  45.13    &  4.95  \\ \hline
    \gr Changen (ours)  &    $P(\I{t+1}|\St{t+1}, \I{t})$      &  \best{34.74}  &  \best{5.41}\\   
  \end{tabular}

\end{table}

\begin{table}[htb]
  \caption{Comparison with other change data generation approaches on the transferability.
  The reported IoU and F$_1$ are obtained by pre-training ChangeStar with synthetic data, fine-tuning it on 5\% LEVIR-CD$^\texttt{train}$ and testing it on LEVIR-CD$^\texttt{test}$, thereby measuring the transferability.
    \label{tab:gen_model_transfer}}
  \centering
  \small
  \tablestyle{8pt}{1.2}
  \begin{tabular}{l|l|cc}
    Method              &  Modeling   &  IoU$\uparrow$  & F$_1$$\uparrow$  \\
    \shline
    SPADE \cite{spade} + GPCM  & \multirow{2}{*}{$P(\I{t}|\St{t})$}        &  -   &  -   \\
    OASIS \cite{oasis} + GPCM  &                                           & 75.9 & 86.3 \\ \hline
    Copy-Paste          & \multirow{3}{*}{$P(\I{t+1}|\St{t+1}, \I{t})$}   & 74.3 & 85.2 \\
    Inpainting          &                                                & 74.6 & 85.5 \\
    \gr Changen (ours)      &           \multicolumn{1}{c|}{(ours)}      & \best{79.3} & \best{88.4} \\   
  \end{tabular}
  \vspace{-1em}
\end{table}

\subsection{Experiments on Change Data Generation}

\mpara{0em}{Setup.}
We train all deep generative models on xView2 pre-disaster and evaluate the quality of synthetic images with FID score \cite{FID} and Inception score (IS) \cite{IS} following common practices.
We do supervised pre-training on each synthetic dataset and then fine-tune the change detection model on 5\% training set of LEVIR-CD \cite{levircd} to measure the transferability.

\mpara{0.em}{Main Result.}
Table~\ref{tab:gen_model} and \ref{tab:gen_model_transfer} presents the comparison with other deep generative models and data-augmentation-based synthesis methods.
For deep generative models, we compare with classical SPADE \cite{spade} and state-of-the-art OASIS \cite{oasis}.
SPADE is hard to generate high-quality images in our settings due to complex global disaster scenarios in xView2, while OASIS yields a FID score of 45.13 and a fine-tuned F$_1$ score of 86.3\%. 
Thanks to our GPCM modeling, our Changen outperforms OASIS both on the image quality and the transferability, yielding a FID score of 34.74 and 88.4\% F$_1$ score.
Besides, OASIS yields stochastic spatial layout due to the absence of temporal information, while our Changen yields more coherent and realistic spatial layout with the pre-event image, as shown in Fig.~\ref{fig:syn_all_m}.
This suggests that the pre-event image guidance is important for the image quality and the transferability if carefully exploited.

For data augmentation-based synthesis approaches, we extend Copy-Paste as \cite{copypaste, chen2021adversarial} in our framework for a fair comparison.
Besides, similar to Copy-Paste (creating objects), we also extend an inpainting methods \cite{inpainting,deepfillv2} (removing objects) as strong baselines, called Inpainting and DeepFillv2.
Since there is only single-temporal data available, above two methods yield bitemporal images with highly consistent background pixels, as shown in Fig.~\ref{fig:syn_all_m}a-b, only generating new and effective positive samples for change detection.
Our Changen produces not only more realistic positive samples but also diverse negative samples via carefully designed masked transition layer to handle the pre-event image guidance, significantly surpassing them by large margins of $\sim$3\% fine-tuned F$_1$ score, respectively.

\begin{table}[htb]
  \caption{\textbf{Ablation study on Masked Transition Layer}, investigating the impact of each component on the image quality and the transferability. 
    \label{tab:ab_mtl}}
  \vspace{-4mm}
  \centering
  \small
  \tablestyle{8pt}{1.2}
  \begin{tabular}{l|cc|cc}
    Method                       & FID$\downarrow$ & IS$\uparrow$ & IoU$\uparrow$ & F$_1$$\uparrow$       \\
    \shline
    Masked Transition Layer      & 34.74   & 5.41                 & 79.3& 88.4        \\\hline
    w/o De-styling               & 37.49   & 5.20                 & 79.1& 88.3        \\
    w/o De-styling and Masking   & 20.38   & 5.00                 & 74.3& 85.2        \\
  \end{tabular}
  \vspace{-4mm}
\end{table}

\mpara{0.em}{Ablation: Masked Transition Layer.}
To yield a accurate change field, \textit{Masking} and \textit{De-styling} operations are two core designs for foreground and background feature weakening, respectively.
Table~\ref{tab:ab_mtl} presents that without \textit{De-styling}, the image quality degenerates slightly (+2.75 FID), but the transferability is hardly impacted (-0.1\% F$_1$).
This suggests that (i) the original style of the pre-event image background region may perturb the image generation, and (ii) the new image style generated from noise can guarantee diverse background regions in the synthesized image for sufficient negative samples of change.

\begin{figure}[ht]
  \centering
   \includegraphics[width=1.\linewidth]{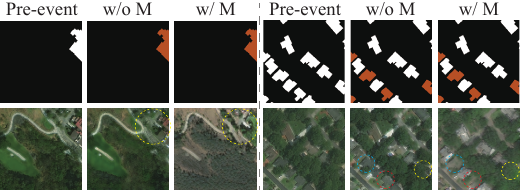}
   \caption{\textbf{Ablation study on the Masked Transition Layer}. ``M'' denotes the masked transition layer.
   Without M, the model trends to copy everything from the pre-event image, failing to remove objects, while with M, the model can successfully remove objects and maintain the spatial layout of the pre-event image.
   }
   \label{fig:ab_m}
   \vspace{-1em}
\end{figure}

Removing \textit{Masking} further, a serious drop of the transferability (-3.2\% F$_1$ score) and an abnormal image quality improvement (-14.36 FID score) are observed because of the trivial solution (``feature leakage'').
This suggests that (i) \textit{Masking} is an indispensable component in the masked transition layer, and (ii) correct positive samples of change is essential to guarantee transferability. 
This is because the trivial solution results in two approximately identical images with different semantic masks, yielding incorrect positive samples of change, while \textit{Masking} explicitly corrects these positive samples, as shown in Fig.~\ref{fig:ab_m}.

\subsection{Experiments on Transferability}

\mpara{0em}{Setup.}
We investigate the transferability using two protocols: (i) zero-shot change evaluation, \ie, the change detector is first pre-trained on synthetic data and then directly tested on the real-world data, and (ii) end-to-end fine-tuning.
The transferability is measured by the model accuracy (\eg, F$_1$ score) for each synthetic data.
We use ChangeStar (FarSeg) with ResNet-18 as the change detector.

\begin{table}[htb]
  \caption{\textbf{Zero-shot Change Evaluation for Object Change Detection} results on LEVIR-CD$^\texttt{all}$ and WHU-CD$^\texttt{all}$.
    ``\Create'' and  ``\Remove'' denote creating objects and removing objects, respectively.
    \label{tab:zero}}
  \vspace{-1em}
  \centering
  \small
  \tablestyle{1.5pt}{1.2}
  \begin{tabular}{l|c|rrrr|rrrr}
                   & Supported        & \multicolumn{4}{c|}{LEVIR-CD$^{\texttt{all}}$} & \multicolumn{4}{c}{WHU-CD$^{\texttt{all}}$}                                             \\
    Pre-train from   & event type       & IoU & F$_1$ & Prec. & Rec. & IoU & F$_1$ & Prec. & Rec. \\
    \shline
    Copy-Paste     & \Create          & 0.7  &  1.4 & 1.9   & 1.2  & 1.9 & 3.8   & 2.4   & 9.0  \\
    Inpainting        & \Remove       & 10.8 & 19.5 & 11.9  & 54.0 & 12.2& 21.8  & 13.3  & 59.3 \\
    OASIS + GPCM & \Create\&\Remove & 39.7  & 56.8 & 45.5  & 75.7 &  12.7 & 22.6 & 13.8  & 61.3 \\\hline
    \gr Changen (ours) & \Create\&\Remove & \best{45.8} & \best{62.8} & \best{49.3}  & \best{86.4} & \best{15.3} & \best{26.6}  & \best{15.7}  & \best{87.1} \\
  \end{tabular}
  \vspace{-1em}
\end{table}

\begin{table}[ht]
  \caption{\textbf{Transfer Learning on LEVIR-CD}. Fine-tuning on LEVIR-CD$^\texttt{train}$ with different training sample ratios and evaluating on LEVIR-CD$^\texttt{test}$.
    ``\Create'' and  ``\Remove'' denote creating objects and removing objects, respectively.
    \label{tab:ft_levir}}
  \vspace{-1em}
  \centering
  \small
  \tablestyle{1.5pt}{1.2}
  \begin{tabular}{l|c|cccc|cccc}
                   & supported        & \multicolumn{4}{c|}{5\%LEVIR-CD$^{\texttt{train}}$ } & \multicolumn{4}{c}{100\%LEVIR-CD$^{\texttt{train}}$}                                             \\
    Pre-train from      & event type       & IoU                                                  & F$_1$                                                & Prec. & Rec. & IoU & F$_1$ & Prec. & Rec. \\
    \shline
    scratch        & -                & 65.4  & 79.1  & 80.3  & 77.9 & 79.4& 88.5  & 89.8  & 87.3 \\
    ImageNet sup.  & -                & 74.3  & 85.2  & 85.9  & 84.5 & 82.7& 90.5  & 91.9  & 89.2 \\
    Copy-Paste     & \Create          & 74.3  & 85.2  & 86.7  & 83.8 & 83.2& 90.8  & \best{92.2}  & 89.5 \\
    Inpainting     & \Remove          & 74.6  & 85.5  & 87.7  & 83.4 & 83.2& 90.8  & 92.0  & 89.6 \\
    DeepFillv2     & \Remove          & 74.8  & 85.6  & 87.5  & 83.7 & 82.7& 90.5  & 91.9  & 89.2 \\\hline
    \gr Changen (ours) & \Create\&\Remove & \best{79.3}  & \best{88.4}  & \best{90.2}  & \best{86.7} & \best{83.7}& \best{91.1}  & 92.0  & \best{90.2} \\
  \end{tabular}
  \vspace{-2.3em}
\end{table}

\mpara{0em}{Zero-shot Change Evaluation.}
Table~\ref{tab:zero} presents the results on the entire LEVIR-CD and WHU-CD \cite{whucd} datasets, respectively, where these two datasets include building object creation and removal events, resulting in two change types.
Pre-training on the synthetic change dataset generated by Copy-Paste yields collapsing zero-shot performance, which suggests that there is an insuperable domain gap between the changes synthesized by Copy-Paste and real-world changes in these two cases at least.
We argue that objects pasted at random positions of the image can be considered as perturbations to this high-dimensional data point, which, if not properly constrained, can move this data point away from the local manifold of the natural image.
Different from Copy-Paste, Inpainting synthesizes post-event images via removing objects and then filling removed regions with its own local statistics.
The perturbations from removal are reduced by the constraints from self-filling, thereby yielding non-collapsing zero-shot performance.

Our Changen significantly outperforms the above two methods by large margins.
Because the perturbations are overcome by the prior from generative models as a constraint.
This prior guarantees that the synthesized image is in training data distribution as much as possible.

\noindent\textbf{Fine-tuning on LEVIR-CD.}
We fine-tune ChangeStar (1$\times$96) for 200 epochs following common practices \cite{levircd, chen2021remote, changestar} (see supplement for training details).
Table~\ref{tab:ft_levir} shows that our Changen yields the synthetic change dataset with superior transferability over other approaches under both 5\% and 100\% training set.
Compared with commonly used ImageNet pre-training, our Changen pre-training significantly improves transferability by 3.2\%/0.6\% F$_1$ under 5\% and full training set, respectively.
This property is very beneficial for those application scenarios where real-world data is limited.
Compared with Copy-Paste, Inpainting, and DeepFillv2, which can be seen the approaches with perfect temporal coherence, our Changen produces more realistic changes and diverse background, \ie, a better trade-off between temporal diversity and temporal coherence.
This suggests that temporal diversity is a key factor in ensuring the transferability of Changen pre-training.

\begin{table}[htb]
  \caption{Comparison with the state-of-the-art change detectors on LEVIR-CD$^{\texttt{test}}$.
  ``R-18'': ResNet-18.
  $\dagger$ indicates that the backbone is pre-trained on ImageNet and then ADE20K \cite{ade20k}.
  $*$ indicates their modified backbone. 
  The amount of floating point operations (Flops) was computed with a float32 tensor of shape [2,256,256,3].
  \label{tab:sota_levir}}
  \vspace{-1em}
  \centering
  \small
  \tablestyle{1.pt}{1.2}
  \begin{tabular}{l|c|c|l|cc}
     Method                   &  Pre-train from        & Backbone        & F$_1\uparrow$         & \#Params. & \#Flops \\
     \shline
     \cb\textit{ConvNet-based}           &       &                 &                       &            &         \\
     FC-Siam-Diff \cite{daudt2018fully}  & -          & -    & 63.0  & 1.3M  & 4.7G     \\
     STANet \cite{levircd}               & ImageNet-1K   & R-18 & 87.3  & 16.9M      &  19.6G    \\
     CDNet+IAug \cite{chen2021adversarial} & ImageNet-1K & R-18 & 89.0  & 14.3M      & -    \\
     BiT \cite{chen2021remote}           &  ImageNet-1K  & R-18 & 89.3  & 11.9M      &   8.6G     \\
     ChangeStar  \cite{changestar}       & ImageNet-1K   & R-18 & 90.2  & 19.3M      &   22.3G    \\
     ChangeStar  \cite{changestar}       & ImageNet-1K   & R-50 & 90.8  & 33.9M      &   29.0G    \\
     ChangeStar  \cite{changestar}       & ImageNet-1K   & RX-101 & 91.2 & 52.5M     &   39.2G    \\\hline
     ChangeStar (1$\times$96)  & ImageNet-1K         & R-18            & 90.5         & 16.4M      &  16.3G  \\
     + self-supervised               &  SeCo-1M \cite{seco}  & R-18       & 89.9         & 16.4M      &  16.3G  \\
     + seg. supervised         & xView2 pre-disaster & R-18     & 90.6         & 16.4M      &  16.3G  \\
     + synthetic data          & \texttt{OASIS-90k}  & R-18     & 90.6         & 16.4M      &  16.3G  \\
     \gr Ours          &  \texttt{Changen-90k}    & R-18            & \multicolumn{2}{l}{91.1\up{0.6}}    & +0  \\
     \multicolumn{2}{l|}{\vb\textit{Transformer-based}}             &        &       &          &             \\ 
     ChangeFormer \cite{changeformer}    & IN-1K,ADE20K$^\dagger$   & MiT-B2$^*$ & 90.4 &       41.0M   &    203.1G         \\
     BiT \cite{chen2021remote,wang2022empirical} & ImageNet-1K & {\scriptsize ViTAEv2-S} &  91.2       &  19.6M   & 15.7G    \\
     \hline
     ChangeStar (1$\times$96)  &  ImageNet-1K  & MiT-B1        &      90.0              &    18.4M          &  16.0G        \\
     \gr Ours                      &  \texttt{Changen-90k}        &  MiT-B1              &   \multicolumn{2}{l}{\best{91.5}\up{1.5}}    &  +0  \\
  \end{tabular}
  \vspace{-2em}
\end{table}

We also compare ChangeStar (1$\times$96) with our Changen pre-training to the state-of-the-art change detection methods and three pre-training baselines in Table~\ref{tab:sota_levir}.
For ConvNet-based change detectors, with R-18 backbone, our Changen pushes ChangeStar further, outperforming other methods with similar complexity, on par with vanilla ChangeStar with RX-101 backbone.
For Transformer-based change detectors, our method with MiT-B1 backbone outperforms BiT with ViTAEv2, setting a new record of 91.5\% F$_1$.
Besides, it is observed that Transformer backbones (\eg, MiT \cite{segformer}) benefits more from our synthetic \texttt{Changen-90k}, improving by 1.5\% F$_1$.
Compared to three strong pre-training baselines (OASIS synthetic data pre-training, segmentation data (xView2 pre-disaster) pre-training, and self-supervised pre-training), our Changen pre-training is superior since it considers temporal dynamics and has zero pretext task gap.

\noindent\textbf{Fine-tuning on S2Looking.}
This dataset is much more challenging because of large geometry offset, radiation differences, tiny change regions, domain gaps between geographic regions \cite{shen2021s2looking}.
We fine-tune ChangeStar (1$\times$96) for 60k iterations (see supplement for training details).
Table~\ref{tab:ft_s2l} shows a consistent observation that Changen pre-training enables the change detector to learn stronger transferable representation.
Changen pre-training outperforms ImageNet pre-training by 3.0\%/0.8\% F$_1$ under 5\% and full training set.
This suggests that the gain from Changen pre-training is hardly affected by the complexity of real-world data, which confirms the robustness of Changen pre-training.

\begin{table}[ht]
  \caption{\textbf{Transfer Learning on S2Looking}. Fine-tuning on S2Looking$^\texttt{train}$ with different training sample ratios and evaluating on S2Looking$^\texttt{test}$.
    ``\Create'' and  ``\Remove'' denote creating objects and removing objects, respectively.
    \label{tab:ft_s2l}}
  \vspace{-1em}
  \centering
  \small
  \tablestyle{1.5pt}{1.2}
  \begin{tabular}{l|c|cccc|cccc}
                   & supported        & \multicolumn{4}{c|}{5\%S2Looking$^{\texttt{train}}$} & \multicolumn{4}{c}{100\%S2Looking$^{\texttt{train}}$}                                             \\
    Pre-train from      & event type       & IoU                                                  & F$_1$                                                & Prec. & Rec. & IoU & F$_1$ & Prec. & Rec. \\
    \shline
    scratch        & -                   & 27.8 & 43.5 & 65.5 & 32.6 & 48.3 & 65.2 & 69.9 & 61.0 \\
    ImageNet sup.  & -                   & 33.2 & 49.9 & 66.1 & 40.0 & 49.6 & 66.3 & 70.9 & 62.2 \\
    Copy-Paste     & \Create             & 34.0 & 50.8 & 68.3 & 40.4 & 50.0 & 66.7 & 69.4 & 64.1 \\
    Inpainting     & \Remove             & 34.0 & 50.8 & \best{69.1} & 40.1 & 49.6 & 66.3 & 69.6 & 63.4 \\
    DeepFillv2     & \Remove             & 34.2 & 51.0  & 66.8       & 41.2 & 50.0 & 66.7 & \best{71.6} & 62.4 \\\hline
    \gr Changen (ours) & \Create\&\Remove    & \best{36.0} & \best{52.9} &65.9 & \best{44.6} & \best{50.5} & \best{67.1} & 70.1 & \best{64.3} \\
  \end{tabular}
  \vspace{-1em}
\end{table}

\begin{table}[ht]
  \caption{Comparison with the state-of-the-art change detectors on the \textbf{S2Looking}$^{\texttt{test}}$ set.
  ``R-18'': ResNet-18.
  The amount of floating point operations (Flops) was computed with a float32 tensor of shape [2,512,512,3] as input.
    \label{tab:sota_s2l}}
  \vspace{-1em}
  \centering
  \small
  \tablestyle{2.5pt}{1.2}
  \begin{tabular}{l|c|ccc|cc}
     Method                             & Backbone        & F$_1\uparrow$         & Prec. & Rec. & \#Params. & \#Flops \\
     \shline
     \cb\textit{ConvNet-based}             &                &                &       &      &            &         \\
     FC-Siam-Diff \cite{daudt2018fully} & -               & 13.1          & 83.2  & 15.7 & 1.3M       & 18.7G   \\
     STANet \cite{levircd}              & R-18            & 45.9          & 38.7  & 56.4 & 16.9M      & 156.7G  \\
     CDNet \cite{chen2021adversarial}   & R-18            & 60.5          & 67.4  & 54.9 & 14.3M      & -       \\
     BiT \cite{chen2021remote}          & R-18            & 61.8          & 72.6  & 53.8 & 11.9M      & 34.7G   \\\hline
     ChangeStar (1$\times$96)           & R-18            & 66.3          & 70.9  & 62.2 & 16.4M      & 65.3G   \\
     \gr + \texttt{Changen-90k}         & R-18            & 67.1          & 70.1  & 64.3 & 16.4M      & 65.3G   \\
                                        &                 & \inc{\bf 0.8}     & \dec{0.8} & \inc{2.1}    &     \multicolumn{2}{c}{+0}       \\
    \multicolumn{2}{l|}{\vb\textit{Transformer-based}}    &                &       &      &            &     \\ 
    ChangeStar (1$\times$96)           & MiT-B1           & 64.3          & 69.3  & 59.9  & 18.4M      & 67.3G   \\
    \gr + \texttt{Changen-90k}         & MiT-B1           & \best{67.9}   & \best{70.3}  & \best{65.7}  & 18.4M   & 67.3G    \\
                                       &                  & \inc{\bf 3.6}     & \inc{1.0} & \inc{5.8}       &  \multicolumn{2}{c}{+0}     \\
  \end{tabular}
  \vspace{-2em}
\end{table}

We further compare ChangeStar (1$\times$96) with our Changen pre-training to the state-of-the-art methods in change detection in Table~\ref{tab:sota_s2l}.
We can see that ChangeStar (1$\times$96) with R-18 or MiT-B1, as our strong baselines pre-trained on ImageNet, have outperformed other methods by large margins.
Nevertheless, our Changen pre-training still surpasses the ImageNet pre-training by 0.8\%/3.6\% F$_1$ for R-18 and MiT-B1 backbones, respectively.

There are two interesting observations:
(i) Transformer-based ChangeStar with ImageNet pre-training is inferior to ConvNet-based when they have similar complexity.
With Changen pre-training, Transformer-based ChangeStar obtains a significant improvement (64.3$\rightarrow$67.9), overtaking ConvNet-based one;
(ii) The improvements from Changen pre-training are mainly from higher recall gains for both ConvNet-based and Transformer-based ChangeStar.

Based on above two observations and consistent results on LEVIR-CD, we argue that (i) large model capacity guarantees the transferability of Changen pre-training.
This is because even though these two models have similar complexity, the model capacity of the transformer is considered to be larger \cite{han2022survey} due to the lack of inductive bias.
(ii) temporal diversity is a key factor in ensuring transferability since it brings rich negative samples that are also important to learn a generalizable change representation, in addition to necessary positive samples.

\begin{figure}[ht]
  \centering
   \includegraphics[width=1.\linewidth]{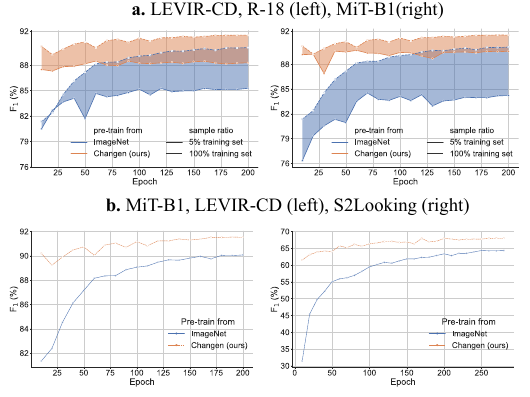}
   \vspace{-1.5em}
   \caption{\textbf{Learning curves} of ChangeStar (1$\times$96) with different pre-training datasets and backbones, respectively.
   \textbf{a.} the accuracy gap is represented by the filled region between solid and dashed lines.
   \textbf{b.} Changen pre-training brings much faster convergence than ImageNet pre-training.
   }
   \label{fig:learning}
   \vspace{-1em}
\end{figure}

\mpara{0em}{Understand Changen Pre-training}.
To understand why the synthetic change dataset yielded by Changen works well on the paradigm of pre-training and fine-tuning, we plot several learning curves in Fig.~\ref{fig:learning} to investigate the behavior of each model.
From Fig.~\ref{fig:learning}a, we can observe that the accuracy gap between the models trained with 5\% and 100\% training set is significantly reduced by our Changen pre-training, compared to ImageNet pre-training.
Besides, our Changen pre-training yields higher average and peak accuracies in different backbones and training sample ratios. 
To our surprise, ChangeStar pre-trained on \texttt{Changen-90k} and fine-tuned on 5\% training set is on-par with it pre-trained on ImageNet and fine-tuned on 100\% when using MiT-B1 backbone.
From Fig.~\ref{fig:learning}b, the model with Changen pre-training has much faster convergence than with ImageNet pre-training.
These results point out that Changen pre-training can find task-specific starting points for deep change detectors, resulting in faster and better convergence.
Transformer-based change detectors can benefit more from Changen pre-training.

\vspace{-1mm}
\section{Conclusion}
\label{sec:conc}
\vspace{-1mm}
We present a scalable multi-temporal remote sensing change data generator to alleviate ``data hungry'' problem for deep change detectors.
Our work enables continuous and controllable multi-temporal change data generation via simulating stochastic change process.
This data generation is easy to extend and scale up via new object properties, and change events.
Our large-scale synthetic change dataset, \texttt{Changen-90k}, is confirmed its superior transferability, making conventional deep change detectors great again, \eg, for the ChangeStar with MiT-B1 backbone, Changen pre-training significantly outperforms ImageNet pre-training and achieves new records on LEVIR-CD and S2Looking datasets.
We also find that temporal diversity is a key factor in ensuring the transferability of Changen pre-training.
These two properties also bring two open problems: the trade-off between fidelity and diversity in image generation, and the trade-off between zero-shot capability and transferability in model generalization, which deserve to be explored in the future.

\section*{Acknowledgements}
This work was supported by National Key Research and Development Program of China under Grant No. 2022YFB3903404, National Natural Science Foundation of China under Grant No. 42325105, 42071350, and LIESMARS Special Research Funding.

{\small
  \bibliographystyle{ieee_fullname}
  \bibliography{egbib}
}

\clearpage

\section*{A. Implementation details}
\subsection*{A.1. Dataset description}

\mpara{0em}{xView2} dataset \cite{gupta2019xbd} is used to benchmark the models for one-to-many semantic change detection \cite{changeos} in the context of the sudden-onset natural disasters.
There are six diaster types of earthquake, wildfire, volcano, storm, flooding, and tsunami in the dataset.
This dataset contains 9,168 image pairs of \texttt{train\&tier3} split, 933 image pairs of \texttt{test} split, and 933 image pairs of \texttt{holdout} split, covering 45,361.79 km$^2$ areas.
Each optical RGB image has a fixed size of 1,024$\times$1,024 pixels.
The images were collected from WorldView-2, WorldView-3, and GeoEye satellites, with varying sub-meter spatial resolutions.
The total of building instances is 850,736.

\mpara{0em}{xView2 pre-diaster}, used in this paper, is the pre-disaster part of xView2 dataset.

\mpara{0.em}{LEVIR-CD} dataset \cite{levircd} consists of 637 bitemporal image pairs, which were collected from the Google Earth platform.
Each image has a fixed size of 1,024$\times$1,024 pixels, with a spatial resolution of 0.5 m.
This dataset provides a total of 31,333 change (building appearing, building disappearing) labels of building instances, but without semantic segmentation masks.
LEVIR-CD dataset is officially split into \texttt{train}, \texttt{val}, and \texttt{test}, three parts of which include 445, 64, and 128 pairs, respectively.

\mpara{0.em}{WHU-CD} dataset \cite{whucd} consists of two aerial images collected in 2012 and
2016, which contains 12,796 and 16,077 building instances respectively. 
Each image has a fixed size of 15,354$\times$32,507 pixels with a spatial resolution of 0.2 m.
The change type is mainly building construction.

\mpara{0.em}{S2Looking} dataset \cite{shen2021s2looking} contains 5,000 image pairs with spatial resolutions from 0.5 to 0.8 m and 65,920 change instances.
The official \texttt{train}, \texttt{val}, and \texttt{test} splits include 3,500, 500, and 1,000 pairs, respectively.
The images were collected from GaoFen, SuperView, and BeiJing-2 satellites of China, which mainly covered globally distributed rural areas.
This dataset features side-looking satellite images, which pose a special yet important challenge that requires the change detector to have sufficient robustness to the registration error and the object geometric offset caused by off-nadir imaging angles.
Each image of this dataset has a fixed size of 1,024$\times$1,024 pixels.

\subsection*{A.2. Implementation details for fine-tuning}

\mpara{0em}{Fine-tuning on LEVIR-CD}.
Random flip, rotate, scale jitter, and cropping into 512$\times$512 are used for training data augmentation.
SGD is used as our optimizer, where the weight decay is 0.0001 and the momentum is 0.9.
The total batch size is 16 and an initially learning rate is 0.03.
We train for 200 epochs on \texttt{train} split, as common practices. 
A ``poly'' learning rate policy ($\gamma$ = 0.9) is applied.

\mpara{0em}{Fine-tuning on S2Looking}.
Random flip, rotate, scale jitter, and cropping into 512$\times$512 are used for training data augmentation.
SGD is used as our optimizer, where the weight decay is 0.0001 and the momentum is 0.9.
The total batch size is 16 and an initially learning rate is 0.03.
We train for 60k iterations on \texttt{train} split.
A ``poly'' learning rate policy ($\gamma$ = 0.9) is applied.

\mpara{0em}{Evaluation metrics}.
F$_1$ score, precision rate (Prec.), and recall rate (Rec.) of change regions are used as evaluation metrics, where F$_1$ score is the main metric.

\section*{B. Scalability of Changen}

\begin{figure*}[ht]
  \centering
   \includegraphics[width=1.\linewidth]{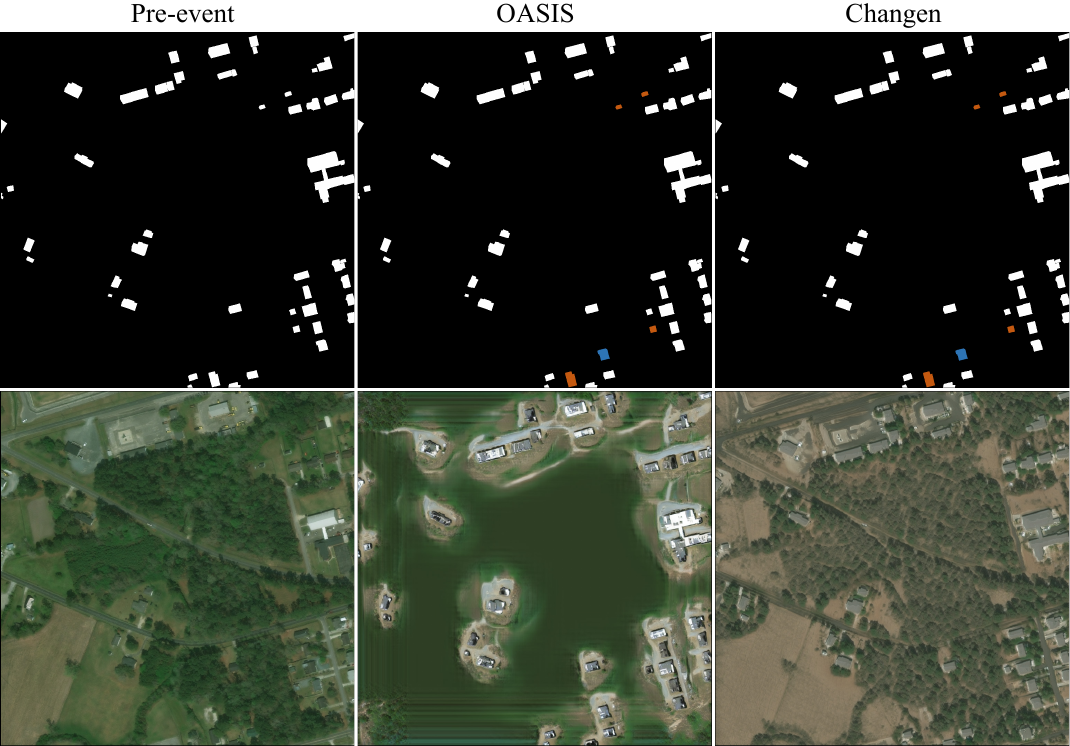}
   \caption{Scaling up the resolution to 1024$\times$1024 pixels.
   }
   \label{fig:resolution1024}
\end{figure*}

\begin{figure*}[ht]
  \centering
   \includegraphics[width=1.\linewidth]{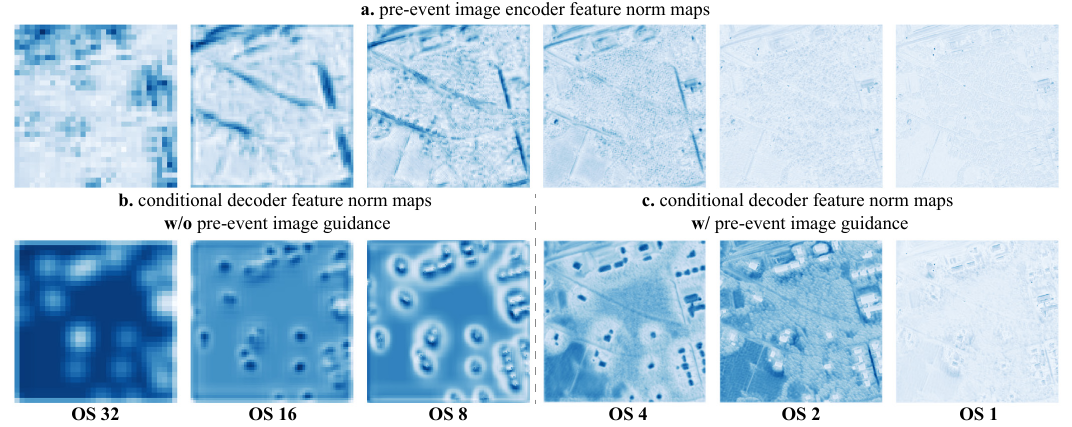}
   \caption{Feature $\ell_2$ norm maps.
   \textbf{a}. the feature norm maps are computed over the pre-event image with Image Encoder of Changen. 
   \textbf{b}. lower-resolution feature norm maps (OS 32, 16, 8) are computed without pre-event image guidance.
   \textbf{c}. higher-resolution feature norm maps (OS 4, 2, 1) are computed with pre-event image guidance.
   ``OS'': output stride.
   }
   \label{fig:featurenorm}
\end{figure*}

\subsection*{B.1. Scaling up Resolution}
Remote sensing images are always of big spatial resolutions beyond 256$\times$256 due to the imaging with high altitudes, \eg, satellite imaging.
Therefore, there is a important requirement that the model trained with 256$\times$256 images can be seamlessly applied to the larger image, \eg, 1024$\times$1024. 
It is easy for discriminative fully convolutional network, however, it is non-trivial for generative models to bridge the resolution gap.
Fig.~\ref{fig:resolution1024} shows the visual results.
OASIS and Changen are both trained with 256$\times$256 images.
OASIS failed to generate realistic image when given a 1,024$\times$1,024 semantic mask.
Obvious artifacts are observed in background region.
Changen still works with this 1024$\times$1024 semantic mask, bridging the resolution gap.

We argue that the main reason why Changen works lies in the pre-event image guidance.
To support our view, we visualize the feature map of each scale via $\ell_2$ norm, as shown in Fig.~\ref{fig:featurenorm}.
We train a variant of Changen, which removes pre-event image guidance in the first three scales (OS 32, 16, 8) and keep pre-event image guidance in the last three scales (OS 4, 2, 1), to investigate the impact of pre-event image guidance.
From Fig.~\ref{fig:featurenorm}b, we can observe that the feature norm map of OS 8 is very similar to the image generated by OASIS in Fig.~\ref{fig:resolution1024}, from the perspective of the background smoothness.
Once applying the pre-event image guidance, observed from Fig.~\ref{fig:featurenorm}c, the feature norm maps look to have more details.
This visual evidence suggests that the pre-event image guidance is the key factor in bridging the resolution gap.

\subsection*{B.2. Scaling up Synthetic Data}

We further scale up the synthetic data from 90k to 1.4M, namely Changen-1.4M, to verify whether data scaling can improve the model performance.
As shown in Table~\ref{tab:syn_data}, our synthetic data volume is at the leading edge.
Scaling up Changen-90k to Changen-1.4M, ChangeStar with MiT-B1 further obtains 0.2\% F$_1$ improvement, achieving 91.7\% F$_1$ on LEVIR-CD.
This suggests that data scaling can improve the model performance.

\begin{table}[htb]
  \caption{Comparison with other synthetic change datasets
    \label{tab:syn_data}}
  \centering
  \small
  \tablestyle{10pt}{1.2}
  \begin{tabular}{l|c|c}
    Dataset name              &  Image size (pixels)   & \#Image pairs  \\
    \shline
    AICD \cite{bourdis2011constrained}       &   800$\times$600   & 1k               \\
    SynCW \cite{kolos2019procedural}       &     3,072$\times$3,072    &      4          \\
    \gr Changen-90k (ours)   & 256$\times$256    &  90k              \\
    \gr Changen-1.4M (ours)  & 256$\times$256    &  1.4M              \\   
  \end{tabular}
\end{table}

\begin{table}[htb]
  \caption{Data scaling results on LEVIR-CD$^{\texttt{test}}$.
  \label{tab:data_scaling_levir}}
  \centering
  \small
  \tablestyle{1.3pt}{1.2}
  \begin{tabular}{l|c|c|l|cc}
     Method                   &  Pre-train from        & Backbone        & F$_1\uparrow$         & \#Params. & \#Flops \\\shline
     ChangeStar (1$\times$96)  &  ImageNet-1K  & MiT-B1        &      90.0              &    18.4M          &  16.0G        \\
     \gr Ours                      &  \texttt{Changen-90k}        &  MiT-B1              &   \multicolumn{2}{l}{91.5\up{1.5}}    &  +0  \\
     \gr Ours                      &  \texttt{Changen}-\texttt{1}.\texttt{4M}        &  MiT-B1              &   \multicolumn{2}{l}{91.7\up{1.7}}    &  +0  \\
  \end{tabular}
\end{table}

\subsection*{C. Comparison with other Pre-training Methods}
Our Changen is a generative model, which is capable of synthesizing multi-temporal change data from \textit{single-temporal segmentation data}.
With synthetic change data (e.g., \texttt{Changen-90k}) pre-training, the change detector gains more on the performance, compared to the commonly used ImageNet-1k supervised pre-training.
Here we investigate the essential effect of Changen pre-training.
The potential performance gain may come from 
(1) less domain gap between pre-train data and downstream data;
(2) semantic segmentation supervision;
(3) zero pretext task gap.
(4) higher-quality synthetic change data.
We discuss these factors next.

\begin{table}[htb]
  \caption{Comparison with other pre-training methods on LEVIR-CD$^{\texttt{test}}$.
  All entries use ResNet-18 as the backbone.
  ``xView2 pre.'': xView2 pre-diaster dataset.
  \label{tab:pretrain_levir}}
  \centering
  \small
  \tablestyle{1.pt}{1.2}
  \begin{tabular}{l|c|c|c}
     Method type                   &  Pre-train method       & Pre-train data                        & F$_1\uparrow$         \\
     \shline
     (a) ChangeStar (1$\times$96)  &  classification          & ImageNet-1K           & 90.5       \\
     (b) + self-supervised         &  SeCo  \cite{seco} & SeCo-1M w/o label                       & 89.9       \\
     (c) + self-supervised         &  MoCov2 \cite{mocov2}& xView2 pre.w/o label                    & 90.4       \\
     (d) + seg. supervised         &  segmentation         & xView2 pre.            & 90.6       \\
     (e) + synthetic data          &  change detection   & \texttt{OASIS-90k}             & 90.6       \\
     \gr (f)  + Ours               &  change detection    & \texttt{Changen-90k}           & 91.1       \\
  \end{tabular}
\end{table}

\mpara{0.5em}{Factor 1}: \textit{less domain gap between pre-train data and downstream data.}
The ImageNet-1k belongs to the natural scenario, which has a large domain gap with the Earth observation scenario.
Can any Earth observation data reduce the domain gap?
The result of Table~\ref{tab:pretrain_levir}(b) gives a negative answer.
With a domain-specific self-supervised method (i.e., SeCo\cite{seco}) and 1 million Sentinel-2 images \cite{drusch2012sentinel}, the transferred change detection performance is instead reduced by 0.6\%F$_1$, compared to ImageNet pre-training.
This is because the spatial resolution of Sentinel-2 optical band is 10 m, while the images of LEVIR-CD has 0.5 m spatial resolution.
The resolution gap is a massive barrier to transfer learning, although the scenario gap has been reduced.

xView2 pre-disaster dataset has sub-meter spatial resolutions close to LEVIR-CD.
However, SeCo requires multi-temporal images as the pre-train data, while xView2 pre-disaster dataset can not meet that since it is single-temporal data.
Thus, we use MoCo v2\cite{mocov2} (the baseline of SeCo) to pre-train the backbone on the xView2 pre-disaster dataset, which yields 90.4\% F$_1$, as Table~\ref{tab:pretrain_levir}(c) presents.
This result somewhat bridges the resolution gap but is still inferior to ImageNet pre-training.

Overall, less domain (e.g., scenario, resolution) gap between pre-train data and downstream data is helpful to transfer the model to the downstream task. However, it is not a primary gain source of our Changen pre-training.

\mpara{0.5em}{Factor 2}: \textit{semantic segmentation supervision.}
In this case, Changen is trained using the xView2 pre-disaster dataset, which is a single-temporal building segmentation dataset.
Therefore, the semantic segmentation supervision provided by this dataset may be a gain source.
We use this segmentation dataset to pre-train the segmentation part of ChangeStar(1$\times$96)\footnote{To whom is not familiar with ChangeStar, ChangeStar can be seen as a segmentation model with a simple change detection head.}. 
This entry yields 90.6\% F$_1$, outperforming ImageNet pre-training by 0.1\% point, as Table~\ref{tab:pretrain_levir}(d) presents.
This result suggests that the semantic segmentation supervision is helpful but it is not a primary gain source of our Changen pre-training.

\mpara{0.5em}{Factor 3}: \textit{zero pretext task gap.}
Our Changen pre-training belongs to synthetic change data pre-training, and pretext task is exactly the change detection.
Therefore, Changen pre-training has zero pretext task gap, which is fundamentally different from self-supervised pre-training, ImageNet pre-training, and segmentation pre-training.
To investigate this factor, we use OASIS, our baseline of the generative model, to synthesize a multi-temporal change dataset (\texttt{OASIS-90k}) as Changen did.
In this way, OASIS pre-training has also zero pretext task gap, which yields 90.6\% F$_1$, as Table~\ref{tab:pretrain_levir}(e) presents.
This result suggests that zero pretext task gap is helpful but it is not a primary gain source of our Changen pre-training.

\mpara{0.5em}{Factor 4}: \textit{higher-quality synthetic change data.}
After ablating other three factors, the last factor matters.
Comparing Table~\ref{tab:pretrain_levir}(e) and (f), all variables are strictly controlled except the synthetic change data from two different generative models, i.e., OASIS and Changen.
Therefore, we argue that the higher quality of the synthetic change data is a primary gain source of our Changen pre-training.
Here, higher quality means better fidelity and diversity of generated images, which is measured by FID (45.13 \textit{vs.} 34.74$_{\rm ours}$, lower is better) and IS (4.95 \textit{vs.} 5.41$_{\rm ours}$, higher is better).

In summary, we find that synthetic change data pre-training is also a promising approach for remote sensing change detection.
This pre-training features less domain gap and zero pretext task gap, where the transferability of pre-trained representation highly depends on the fidelity and diversity of generated images. 
Zero pretext task gap means that the model pre-trained in this way has zero-shot prediction capability, which other pre-training methods (above discussed) cannot achieve.

\end{document}